\documentclass{article}

\usepackage[final,nonatbib]{neurips_2024}

\usepackage[utf8]{inputenc} 
\usepackage[T1]{fontenc}    
\usepackage{hyperref}       
\usepackage{url}            
\usepackage{booktabs}       
\usepackage{amsfonts}       
\usepackage{nicefrac}       
\usepackage{microtype}      
\usepackage{xcolor}         

\usepackage{graphicx}
\usepackage{amsmath}
\usepackage[linesnumbered,ruled,vlined]{algorithm2e}
\usepackage{amssymb}
\usepackage{ragged2e}
\usepackage{lipsum} 
\usepackage{makecell}

\title{Localize, Understand, Collaborate:\\
Semantic-Aware Dragging via Intention Reasoner}

\author{
Xing Cui$^1$, Peipei Li$^1$$^{\star}$, Zekun Li$^2$, Xuannan Liu$^1$, Yueying Zou$^1$, Zhaofeng He$^1$ \\
    $^{1}$Beijing University of Posts and Telecommunications \\
    $^{2}$University of California, Santa Barbara \\
    \{cuixing, lipeipei, liuxuannan, zouyueying2001, zhaofenghe\}@bupt.edu.cn\\
    zekunli@cs.ucsb.edu\\
}

\newcommand{\eps}{{\epsilon}}

\begin{document}

\maketitle
\let\thefootnote\relax\footnotetext{$^{\star}$Corresponding author}

\begin{figure*}[!ht]
  \centering
    \includegraphics[width=1.0\linewidth]{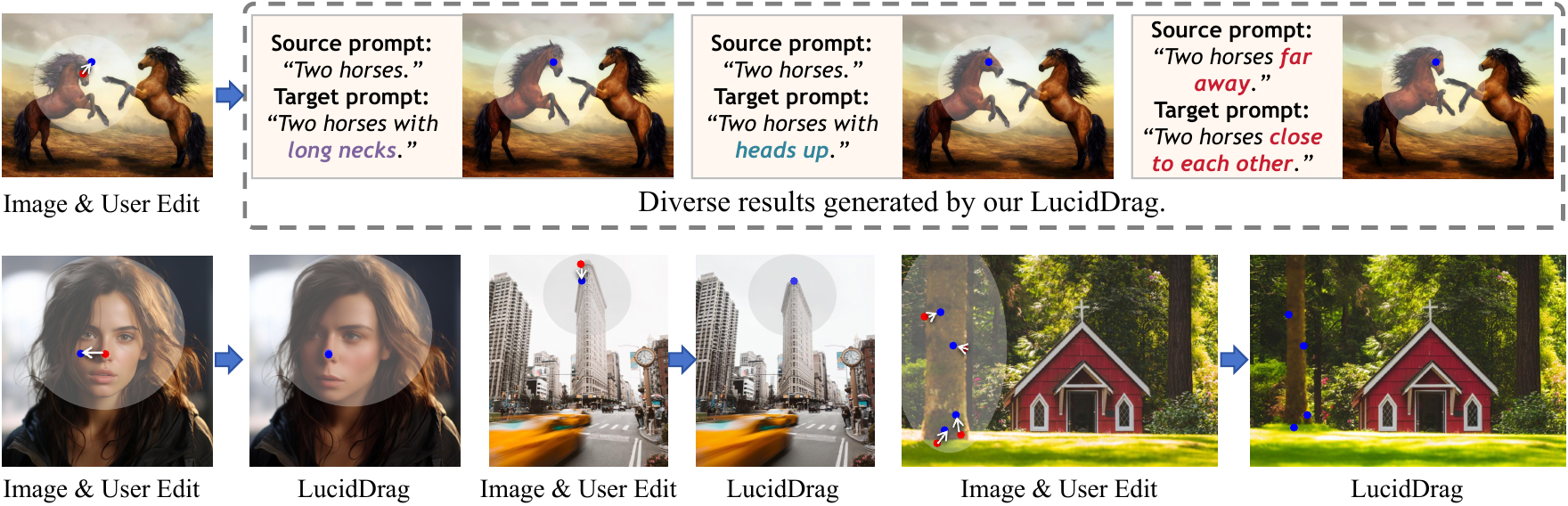}
    \caption{
    Given an input image, the user draws a mask specifying the editable region and clicks dragging points (handle points (\textcolor{red}{red}) and target points (\textcolor{blue}{blue})). 
    Our LucidDrag considers the ill-posed nature of drag-based editing and can produce diverse results (the first row). Besides,  it achieves outstanding performance in editing accuracy and image fidelity (the second row). 
    }
    \label{fig:intro-image}
\end{figure*}

\begin{abstract}
Flexible and accurate drag-based editing is a challenging task that has recently garnered significant attention. Current methods typically model this problem as automatically learning "how to drag" through point dragging and often produce one deterministic estimation, which presents two key limitations: 1) Overlooking the inherently ill-posed nature of drag-based editing, where multiple results may correspond to a given input, as illustrated in Fig.~\ref{fig:intro-image}; 2) Ignoring the constraint of image quality, which may lead to unexpected distortion.
To alleviate this, we propose LucidDrag, which shifts the focus from "how to drag" to "what-then-how" paradigm.  LucidDrag comprises an intention reasoner and a collaborative guidance sampling mechanism. The former infers several optimal editing strategies, identifying what content and what semantic direction to be edited. Based on the former, the latter addresses "how to drag" by collaboratively integrating existing editing guidance with the newly proposed semantic guidance and quality guidance.
Specifically, semantic guidance is derived by establishing a semantic editing direction based on reasoned intentions, while quality guidance is achieved through classifier guidance using an image fidelity discriminator.
Both qualitative and quantitative comparisons demonstrate the superiority of LucidDrag over previous methods. Code is available at: \url{https://github.com/cuixing100876/LucidDrag-NeurIPS2024}.

\end{abstract}

\section{Introduction}
The impressive success of diffusion models~\cite{ho2020denoising,dhariwal2021diffusion,rombach2022high} has promoted the advancements in the field of image editing~\cite{hertz2022prompt,zhang2023adding,gal2022image}. The conventional paradigms for editing conditions typically encompass text~\cite{dong2023prompt,nam2023contrastive,hertz2023delta}, instruction~\cite{brooks2023instructpix2pix,zhang2023hive,zhang2024magicbrush,fu2024guiding}, or image~\cite{Yang_2023_ICCV,Wang_2023_ICCV,Gu_2023_ICCV}. 
However, these conditions prove inadequate in effectively communicating specific image aspects, such as shape and location~\cite{epstein2023diffusion}.

To address this, recent studies ~\cite{endo2022user,pan2023drag,shi2023dragdiffusion} define a new task called drag-based editing, which incorporates dragging points as conditions. These studies specifically regard drag-based editing as the problem of "how to drag" and tackle it by designing an editing guidance loss, enabling the model to implicitly learn the appropriate solutions.
Despite their considerable success~\cite{pan2023drag, mou2023dragondiffusion, mou2024diffeditor}, these methods have two limitations: 
\textit{\textbf{Firstly, they neglect the inherent ambiguity of semantic intention. }} 
Drag-based editing is an ill-posed problem. As depicted in Fig.~\ref{fig:intro-image}, the drag points starting from the horse's head and ending at its upper right can indicate various semantic intentions, such as "make the neck longer," "raise the head," or "bring the two horses closer."
However, existing methods mainly follow point dragging principles~\cite{shi2023dragdiffusion, mou2023dragondiffusion}, focusing on positional movement by constraining feature correlation between the source and target points. This current position optimization strategy inherently overlooks semantic diversity, making it challenging to generate images with precise semantic perception.
\textit{\textbf{Secondly, they overlook the preservation of the overall image quality.}} Current methods prioritize editing accuracy while neglects the overall image quality. Some methods~\cite{mou2023dragondiffusion, mou2024diffeditor} utilize score-based classifier guidance for image editing, which can cause mismatches between the distribution of the edited image and the input image, compromising image fidelity.

In this study, we divide the drag-based editing task into two steps. We introduce a preliminary step, "what to drag", to determine the specific content and semantic direction for editing before addressing "how to drag". That is, we shift the focus from "how to drag" to a paradigm of  "what-then-how". 
As shown in the first row of images in Fig.~\ref{fig:intro-image}, before editing, we need to determine what we are going to edit, such as the horse's head, and the semantic strategy for editing it towards the upper right. For example, we could make the horse lift its head, elongate its neck, or shorten the distance between the two horses. With the "what to drag" information established, we can then proceed to address "how to drag".
To achieve this, we construct an intention reasoner that integrates a Large Language-Vision Model (LVLM) and a Large Language Model (LLM) to deduce possible intentions. 
As illustrated in Fig.~\ref{fig:intro-image}, given an input image and drag points that begin at the horse's face and end in its upper-right, the intention reasoner acts as an AI agent to infer potential intentions, subsequently providing corresponding source and target prompts. 
Once the intention is determined, we then address "how to drag" by injecting the reasoned semantic information into the model by developing collaborative guidance sampling, 
which integrates editing guidance with the proposed semantic guidance and quality guidance. 
Specifically, the semantic guidance is derived from the source and target prompts determined by the intention reasoner. The asymmetric prompts establish a clear semantic editing direction towards the target intention.
Additionally, a discriminator is employed as the score function to provide quality guidance. The quality gradient is generated based on image fidelity and incorporated into the model via the classifier guidance mechanism.
For the editing guidance, we follow previous work~\cite{mou2023dragondiffusion} to maximize the feature correspondence between the source and target positions.

LucidDrag is an intuitive framework for "what-then-how" drag-based editing, demonstrating outstanding performance in terms of semantic perception ability, diversity, and editing quality. LucidDrag enjoys several attractive attributes:
\textbf{\textit{Firstly, clear, diverse, and reasonable semantic intentions.}} Our innovative method employs LVLM and LLM to explicitly deduce the intention by localizing the drag points and reasoning several probable intentions. By explicitly incorporating semantics, we enhance semantic perception and offer diverse editing modes, enriching the variety of outcomes.
\textbf{\textit{Secondly, enhanced overall generation quality.}} By introducing collaborative guidance sampling, we significantly promote the generation quality of drag-based editing. Specifically, we achieve diverse and accurate image editing by introducing an additional semantic editing direction through semantic guidance. Additionally, we maintain better image quality by explicitly constraining the image distribution using quality guidance.
In summary, our main contributions are:
\begin{itemize}
    \item We propose a new "what-then-how" paradigm for drag-based editing and introduce LucidDrag. To address the "what" problem, we present an intention reasoner, which employs LVLM and LLM to determine what content and semantic direction should be edited.
    \item We then use the inferred semantic results of "what" to guide "how", enhancing editing accuracy and overall image quality. Additionally, a quality discriminator is also employed to provide quality gradients via score-based classifier guidance. This quality guidance, combined with editing and semantic guidance from "what," improves the precision and fidelity of the results.
    \item We present quantitative and qualitative results demonstrating the applicability and superiority of our method, in terms of editing accuracy, fidelity, and diversity.

\end{itemize}

\section{Related Work}
\subsection{Diffusion Models}
Diffusion model~\cite{sohl2015deep, graikos2022diffusion, song2020score} aims to estimate the noise $\eps_t$ added to the image $z_t = \alpha_t x + \sigma_t \eps_t$ , where $\alpha_t$ and $\sigma_t$ are non-learned parameters. The training loss is to minimize the distance between the added noise and the estimated noise:
\begin{equation}
    L(\theta) = \mathbb{E}_{t \sim \mathcal{U}(1, T), \eps_t \sim \mathcal{N}(0, {I})} || \eps_t - \eps_\theta(z_t; t, y) ||_2^2,
\end{equation}
where $t$ refers to the time step, $\eps_t$ is the ground-truth noise, $y$ is an additional condition.
Diffusion models can be regarded as score-based models~\cite{song2020score}. In this context,  $\eps_\theta$ serves as an approximation of the score function for the noisy marginal distributions: $\eps_\theta(z_t) \approx \nabla_{z_t} \log p(z_t)$.

we can sample images given conditioning $y$ by starting from a random noise $z_T \sim \mathcal{N}(0, I)$, and then alternating between estimating the noise component and sampling $z_{t-1}$. The noise is estimated as:
\begin{equation}
\label{eq:pred_noise}
\hat\eps_t = \eps_\theta(z_t; t, y).
\end{equation}
The sampling could be based on DDPM \cite{graikos2022diffusion} or DDIM \cite{song2020denoising}. In this paper, we utilize DDIM, which denoising $z_t$ to a previous step $z_{t-1}$ with a deterministic process:
\begin{equation}
\label{eq:ddim}
\resizebox{0.43\textwidth}{!}{
    $z_{t-1}=\sqrt{\frac{\alpha_{t-1}}{\alpha_{t}}}z_{t}+\left(\sqrt{\frac{1}{\alpha_{t-1}}-1}-\sqrt{\frac{1}{\alpha_{t}}-1}\right)\cdot\hat\eps_t$.
    }
\end{equation}

\subsection{Classifier Guidance}
Classifier guidance moves the sampling process towards images that are more likely according to the classifier \cite{dhariwal2021diffusion}. As a powerful conditional sampling strategy, it has been used in various tasks, including generating diverse results~\cite{corso2023particle}, refining generative process~\cite{kim2023refining,yoon2024censored} and image editing~\cite{mou2023dragondiffusion,mou2024diffeditor}.
Specifically, it combines the unconditional score function for $p(z_t)$ with a classifier $p(y | z_t)$ to produce samples from $p(z_t | y) \propto p(y|z_t)p(z_t)$ ~\cite{dhariwal2021diffusion, song2020score}.
Classifier Guidance requires access to a labeled dataset and the training of a noise-dependent classifier $p(y|z_t)$ which can be differentiated concerning the noisy image $z_t$. During the sampling process, classifier guidance can be incorporated as follows:
\begin{equation}
   \nabla_{{z}_t} \log q({z}_t | {y}) \propto \nabla_{{z}_t} \log q({z}_t)+\nabla_{{z}_t} \log q({y}|{z}_t), 
\end{equation}
The first term is the original diffusion denoiser, and the second term refers to the conditional gradient produced by an energy function $g\left({z}_t; t, {y}\right)=q\left({y}|{z}_t \right)$. Thereby, we apply classifier guidance by modifying $\hat\eps_t$:
\begin{equation}
\hat\eps_t = \eps_\theta(z_t; t, y) + \eta  \nabla_{z_t} \log g\left({z}_t; t, {y}\right), 
\label{eq_classifierguid}
\end{equation}
where $\eta $ is an additional parameter parameter that modulates the strength of the guidance.

\subsection{Visual Programming}
As LLMs~\cite{chatgpt2022,chiang2023vicuna,touvron2023llama} and LVLMs~\cite{zhu2023minigpt,dai2024instructblip,yuan2023osprey} demonstrated remarkable emergency abilities~\cite{achiam2023gpt,wei2022chain,liu2024fka, liu2024mmfakebench}, researches~\cite{gani2023llm,lian2023llm,vashishtha2024chaining} explore to leverage them for planning and reasoning in multi-modal image generation. For example, ChatEdit~\cite{cui2023chatedit} utilizes pre-trained language models to track the user intents.
Some approaches~\cite{gani2023llm,lian2023llm,vashishtha2024chaining} augment the original prompt through paraphrasing. 
Recent, visual programmer methods~\cite{gupta2023visual,han2023image} translate complex input prompts into programmatic operations and data flows.
Despite the effectiveness of these strategies in augmenting the input text instructions, they overlook the capacity to reason within visual-modal instructions, such as the dragging points in drag editing tasks. In contrast, our LucidDrag addresses this gap by integrating LVLM and LLM.

\subsection{Image Editing}
Image editing aims to manipulate an image according to specific conditions. Prior methodologies~\cite{li2019global,li2019m2fpa,li2020hierarchical} can only manipulate specific attributes. The prevailing approaches have primarily relied on text conditions. For example, \cite{cui2023chatedit,alaluf2022hyperstyle} manipulate images in the GAN~\cite{karras2019style} latent space by learning an edit direction. Motivated by the success of the diffusion model~\cite{ho2020denoising,wang2024stablegarment}, state-of-the-art methods extend their exploration into diffusion-based image editing by exploring the initial noise~\cite{cui2023instastyle,zhao2023null,mao2023guided}, attention maps~\cite{chen2024training,hertz2022prompt,xie2023boxdiff,epstein2023diffusion}, or prompts~\cite{li2023pluralistic,mokady2023null,dong2023prompt,brooks2023instructpix2pix,teng2024exploring}. 
Recently, DragGAN~\cite{pan2023drag} explores a novel editing scheme that drags any points of the image to reach target points with the help of StyleGAN~\cite{karras2019style} latent space. FreeDrag~\cite{ling2023freedrag} improves point tracking by introducing adaptive feature updating and backtracking. 
Readout Guidance~\cite{luo2023readout} solves the challenging task by leveraging the video dataset. 
The following works~
\cite{mou2023dragondiffusion,mou2024diffeditor} utilize the feature correspondence to direct the editing process. 
Nevertheless, they only use the drag points as a control, which is insufficient due to the potential diversity in semantic intentions. 
In contrast, our approach introduces an intention reasoner to achieve semantic-aware editing.  The intention reasoner can reduce cognitive load, handle vague requests, and discover potential needs. Additionally, it generates precise descriptions automatically, ensuring accurate and consistent manipulations.

\begin{figure*}
  \centering
    \includegraphics[width=\linewidth]{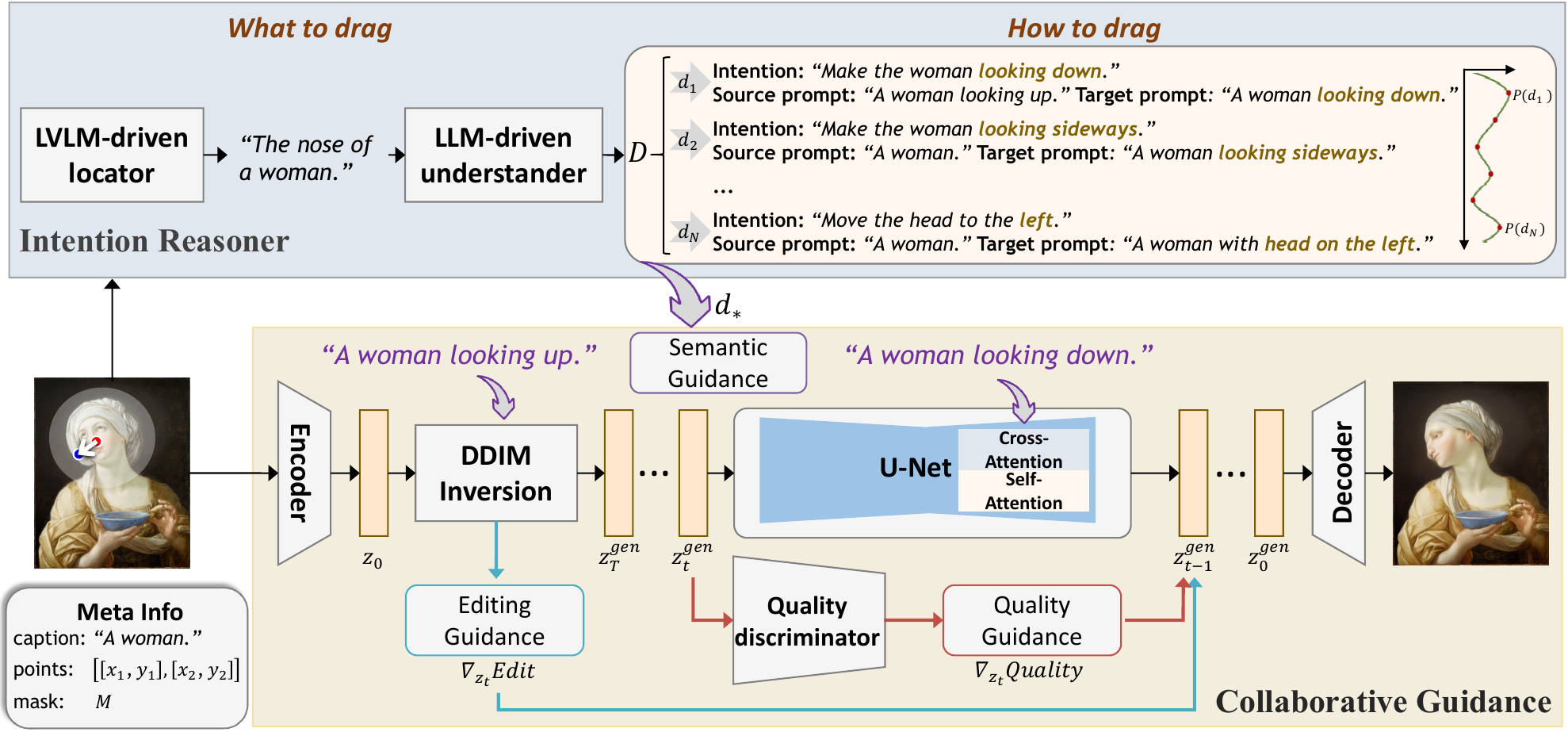}
    \caption{Overview of LucidDrag. LucidDrag comprises two main components: an intention reasoner and a collaborative guidance sampling mechanism. \textit{Intention Reasoner} leverages an LVLM and an LLM to reason $N$ possible semantic intentions. 
    \textit{Collaborative Guidance Sampling} facilitates semantic-aware editing by collaborating editing guidance with semantic guidance and quality guidance.}
    \label{fig:main-diagram}
\end{figure*}

\section{Method}
In this section, we introduce LucidDrag, a unified framework for Drag Manipulation via \textbf{{L}}ocaling, \textbf{{U}}nderstanding, \textbf{C}ollaborate Gu\textbf{{id}}ing. Within our framework, image drag editing is decomposed into two stages. Firstly, the \textbf{\textit{Intention Reasoner}} translates the user-drag points into potential semantic intentions and generates corresponding source prompt and target prompt, thus solving the problem of "what to edit". Then, the \textbf{\textit{Collaborative Guidance Sampling}} is designed to facilitate image editing. Concretely, prompts generated from the intention reasoner are utilized in the DDIM Inversion and Diffusion U-Net, providing semantic guidance for the generation of semantically controllable results. Besides, LucidDrag employs a discriminator to provide quality guidance for improved image fidelity. The semantic guidance and the quality guidance collaborate with the original editing guidance, offering a novel perspective of "how to edit". We elaborate on more details of our techniques below.

\subsection{Intention Reasoner}
As depicted in Fig.~\ref{fig:main-diagram}, the first stage of our LucidDrag is the intention reasoner, which bridges the gap between the input point condition and semantic intention. This intention reasoner consists of two key components, \textit{i.e.}, an LVLM locator that identifies the interested position, and an LLM understander that interprets input conditions into semantic intentions.

\paragraph{LVLM-driven loactor}
Given an input image, it may contain various objects situated at different positions. To accurately identify the objects of interest, we employ an off-the-shelf pre-trained large vision-language model (LVLM) Osprey~\cite{yuan2023osprey}. Osprey is trained with fine-grained mask regions, enabling it to comprehend images at the pixel level. With the input image$x$, we instruct the Osprey model with drag points $P$ to generate a descriptive representation of the objects of interest $O$, \textit{i.e.}, $O=LVLM(x, P)$. As shown in Fig.~\ref{fig:main-diagram}, $O$="\textit{The nose of a woman}", which subsequently serves as input to prompt the LLM to understand the condition and reason potential intentions. 

\paragraph{LLM-driven reasoner} As shown in Fig.~\ref{fig:main-diagram}, each point condition may encompass various semantic intentions. For instance, it may represent non-rigid manipulation, such as "looking down", while it may also represent rigid manipulation, such as "moving to the left".
Our LLM-driven reasoner is designed to discern potential semantic intentions to facilitate semantic-aware drag-based editing. 
We leverage the capabilities of the large language model, GPT 3.5~\cite{chatgpt2022}, acting as an AI agent to reason the possible intentions. We take the generated description of the object of interest $O$, the original image caption $C$, and drag points $P$ as input. Then, we prompt the LLM with in-context examples to generate $N$ possible intentions, \textit{i.e.} $D=LLM(O,C,P)$, where each output sample can have different intentions or levels of complexity. Specifically, $D=\left \{ \left ( d_j, P\left ( d_{j} \right ) \right ) \right \}_{j=1}^{N}$, where $d_j=\left \{i_{j},p_{j}^{s}, p_{j}^{t}\right\}$ is the generated text output. $i_{j}$, $p_{j}^{s}$, $p_{j}^{t}$ represents the predicted intention, the predicted description of the source image (source prompt), and the predicted description of the target image (target prompt), respectively. 
$P\left ( d_{j} \right )$ is the corresponding confidence probabilities of the $j\text{-}th$ generate text output. 
Finally, we can select $n$ outputs $\left \{  d_{s^*} \right \} _{s=1}^{n}$ by sampling based on the confidence probabilities. The confidence probability reflects the quality of the output. A higher confidence probability indicates that the intention is more reasonable, leading to better editing results.

\begin{equation}
    \left \{  d_{s^*} \right \} _{s=1}^{n}=\underset{d_{i}\in D}{argmax}\left ( P\left ( d_{i} \right ), n \right ).
\end{equation}

\subsection{Collaborative Guidance Sampling}
As shown in Fig.~\ref{fig:main-diagram}, the objective of collaborative guidance sampling is to modify the intended content while ensuring the preservation of irrelevant components. 
The input image is inverted to noise $z_T^{gen}$ through DDIM Inversion~\cite{song2020denoising,mokady2023null}. During the inversion process, the intermediate noise $z_t^{gud}$, along with the corresponding key $k_t^{gud}$, and value $v_t^{gud}$ of the self-attention layer, are recorded in the memory bank, which serves in guiding subsequent generation process.
Subsequently, we generate the edited images employing collaborative guidance sampling which incorporates three fundamental components: semantic guidance, quality guidance, and editing guidance. 
Each of these components contributes to the overall editing process distinctively, thereby ensuring a balanced and comprehensive approach to image editing.

\paragraph{Semantic guidance}
As textual conditions can convey semantic information, we leverage the source and target prompts generated by the intention reasoner to facilitate semantic-aware dragging. 
Specifically, during the inversion process, we employ the source prompt to transform the input image into its corresponding noise by iterating DDIM inversion, \textit{i.e.}, ${z}_{t+1}^{gud}=\mathit{DDIM\_inversion}({z}_{t}^{gud},p_{*}^{s})$.
During the sampling process, we utilize the target prompt to generate the target image, \textit{i.e.}, ${z}_{t-1}^{gen}=\mathit{DDIM} ({z}_{t}^{gen},p_{*}^{t})$.
As there exists a divergence between the source and target prompts, the asymmetric textual condition introduces a distinct editing direction that is oriented toward the target image, thereby offering semantic guidance. This differentiation facilitates a semantically guided editing process, enhancing the semantic coherence of the edited image.

\paragraph{Quality guidance}
As shown in Fig.~\ref{fig:main-diagram}, we design quality guidance to ensure the quality of the generated image. A discriminator is trained to distinguish between high-fidelity images and low-fidelity images at any step $t$. In particular, given a real image and its corresponding text description $y$, we utilize a stochastic process to simulate potential points of drag and their respective directions. Then, we generate images using DragonDiffusion~\cite{mou2023dragondiffusion}. 
The images with an aesthetic score~\cite{schuhmann2022laion} below $5$, are classified as low-fidelity.  Their corresponding real images are considered high-fidelity.
The selected low-fidelity images, along with the high-fidelity images, constitute the training dataset of the discriminator. Finally, the training dataset comprises a total of 10,000 high-fidelity images and 10,000 low-fidelity images.  

As the intermediate representation of the diffusion U-Net captures semantic information of the input image~\cite{kwon2023diffusion}, we utilize this hidden representation to evaluate image quality. The discriminator comprises the down block and middle block of the Diffusion U-Net to capture the semantic information, followed by a linear classifier layer. During training, we froze the down blocks which are initialized with Stable Diffusion v2.1-base~\cite{rombach2022high}. The middle block and the prediction layer are fine-tuned to classify images as real or fake. The conditional discriminator $d(X_t| y; t)$ is trained by minimizing the canonical discrimination loss:
\begin{equation}
\begin{aligned}
\mathcal{L} =  \mathbb{E}_{y, t}\big[&\mathbb{E}_{z_t \sim p(z_t | y)}[-\log d(z_t| y; t)] \\
+ & \mathbb{E}_{z_t \sim q(z_t | y)}[-\log (1-d(z_t| y; t))]\big].
\end{aligned}
\label{eq:loss}
\end{equation}

The energy function to constrain the image quality is defined as~\cite{goodfellow2020generative,kim2023refining}:
\begin{equation}
    g_{quality} = \frac{p(z_t | y)}{q(z_t | y)}\approx \frac{d^*(z_t | y; t)}{1 - d^*(z_t | y; t)}.
    \label{eq_gquality}
\end{equation}

\paragraph{Editing guidance}
Following DragonDiffusion~\cite{mou2023dragondiffusion}, we extract intermediate features ${F}_{t}^{gen}$ and ${F}_{t}^{gud}$ from ${z}_{t}^{gen}$ and ${z}_{t}^{gud}$ via UNet denoiser $\epsilon_{\boldsymbol{\theta}}$ respectively. The energy function is built by calculating the correspondence between ${F}_{t}^{gen}$ and ${F}_{t}^{gud}$. 
The editing guidance includes editing target contents and preserving unrelated regions.
We denote the original content position, target content position, and the unrelated content position as binary masks, \textit{i.e.}, ${m}^{orig}$, ${m}^{tar}$,  and ${m}^{share}$, respectively. 
The energy function to drag the content is defined as:
\begin{equation}
\label{eq_gdrag}
    g_{drag} = \frac{1}{\alpha+\beta \cdot \mathcal{S}({F}_{t}^{gen}, {m}^{tar}, {F}_{t}^{gud}, {m}^{orig})},
\end{equation}
where $\mathcal{S}({F}_{t}^{gen}, {m}^{tar}, {F}_{t}^{gud}, {m}^{orig})$ calculates the similarity between the two regions of ${F}_{t}^{gen}$ and ${F}_{t}^{gud}$. Similarly, the energy function to preserve the unrelated region is defined as:
\begin{equation}
\label{eq_gcontent}
    g_{content} =  \frac{1}{\alpha+\beta \cdot \mathcal{S}_{local}({F}_{t}^{gen}, {m}^{share}, {F}_{t}^{gud}, {m}^{share})}.
\end{equation}

The final editing energy function is defined as:
\begin{equation}
\label{eq_edit}
    g_{edit} = w_{e}\cdot  g_{drag}+ w_{c}\cdot g_{content},
\end{equation}
where $w_{e}$ and $w_{c}$ are hyper-parameters to balance these guidance terms.
Finally, the editing guidance collaborates with the quality guidance during sampling:

\begin{equation}
    g\left({z}_t; t, {y}\right)=g_{edit}+g_{guality}.
\label{eq_collabguid}
\end{equation}

Additionally, following~\cite{cao2023masactrl}, to ensure content consistency between the edited image and the input images, we replace the keys and values within the self-attention module of the UNet decoder with those retrieved from the memory bank.

\section{Experiments}
\subsection{Implementation Details}\label{sec:experiments_details}
To train the quality discriminator, we employ the Adam optimizer with a learning rate of 1e-4. We set the training epochs as 100 and the batch size as 128. 
For the denoising process, we adopt Stable Diffusion~\cite{rombach2022high} as the base model. During sampling, the number of denoising steps is set to $T=50$ with a classifier-free guidance of $5$. The energy weights for $g_{quality}$, $g_{drag}$ and $g_{content}$ are set to $1e-3$, $4e-4$ and $4e-4$, respectively.  The training of the discriminator can be conducted on a NVIDIA V100 GPU and the inference can be conducted on a NVIDIA GeForce RTX 3090 GPU.

\subsection{Comparisons}
\begin{figure*}
  \centering
    \includegraphics[width=\linewidth]{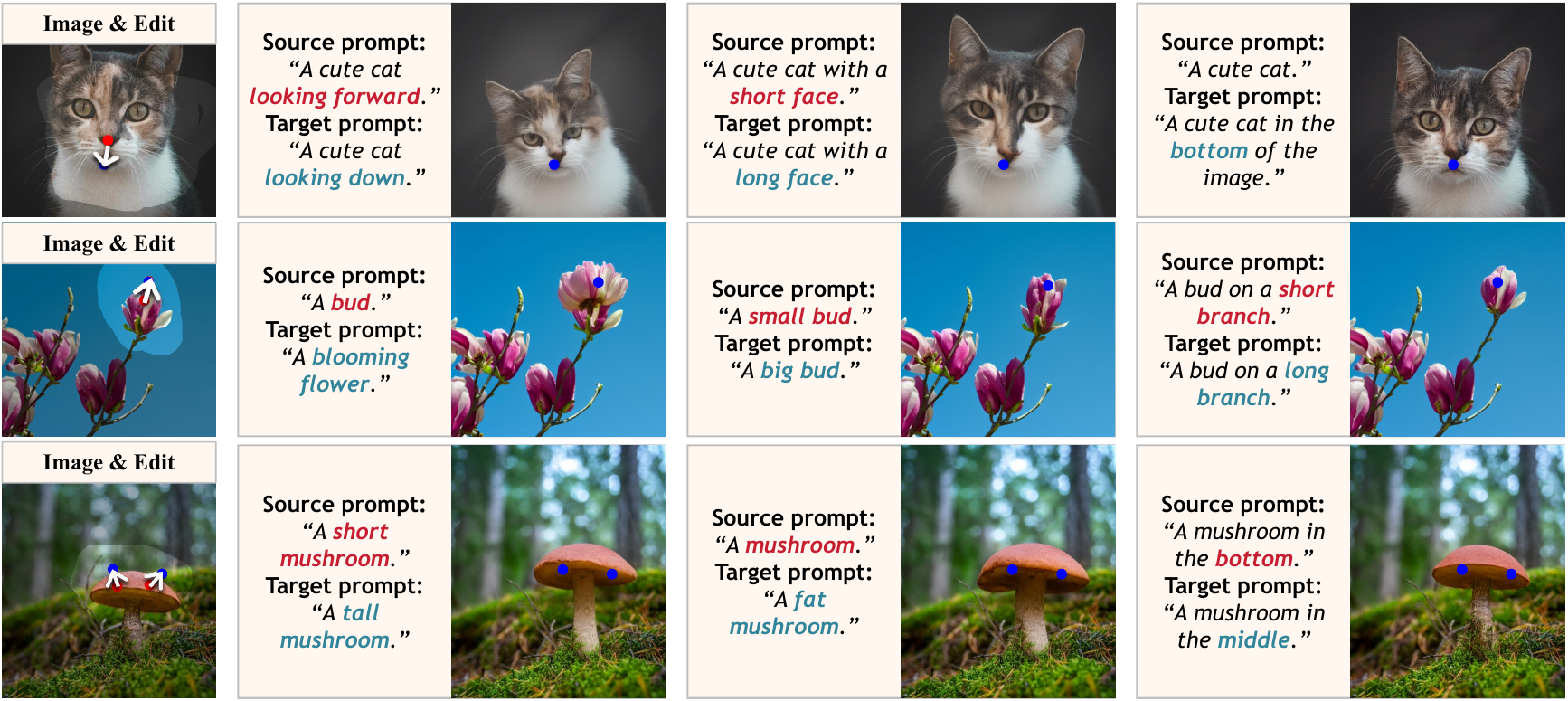}
    \caption{LucidDrag allows generating diverse results conforming to the intention.}
    \label{fig:main-diversity}
\end{figure*}

\paragraph{Semantic-aware dragging} Since our LucidDrag effectively discerns potential intentions, we first evaluate its semantic-awareness ability. Specifically, given an input image and corresponding dragging conditions, we sample several intentions and obtain corresponding results. The results are shown in Fig.~\ref{fig:main-diversity}. On the one hand, the intention reasoner deduces reasonable intentions that align with the input dragging points and generate semantic-aware images, demonstrating both an enhanced understanding of semantic intentions and increased diversity. On the other hand, our method can generate high-fidelity images aligned with the input prompts, improving the quality of the results.

\begin{figure*}
  \centering
    \includegraphics[width=\linewidth]{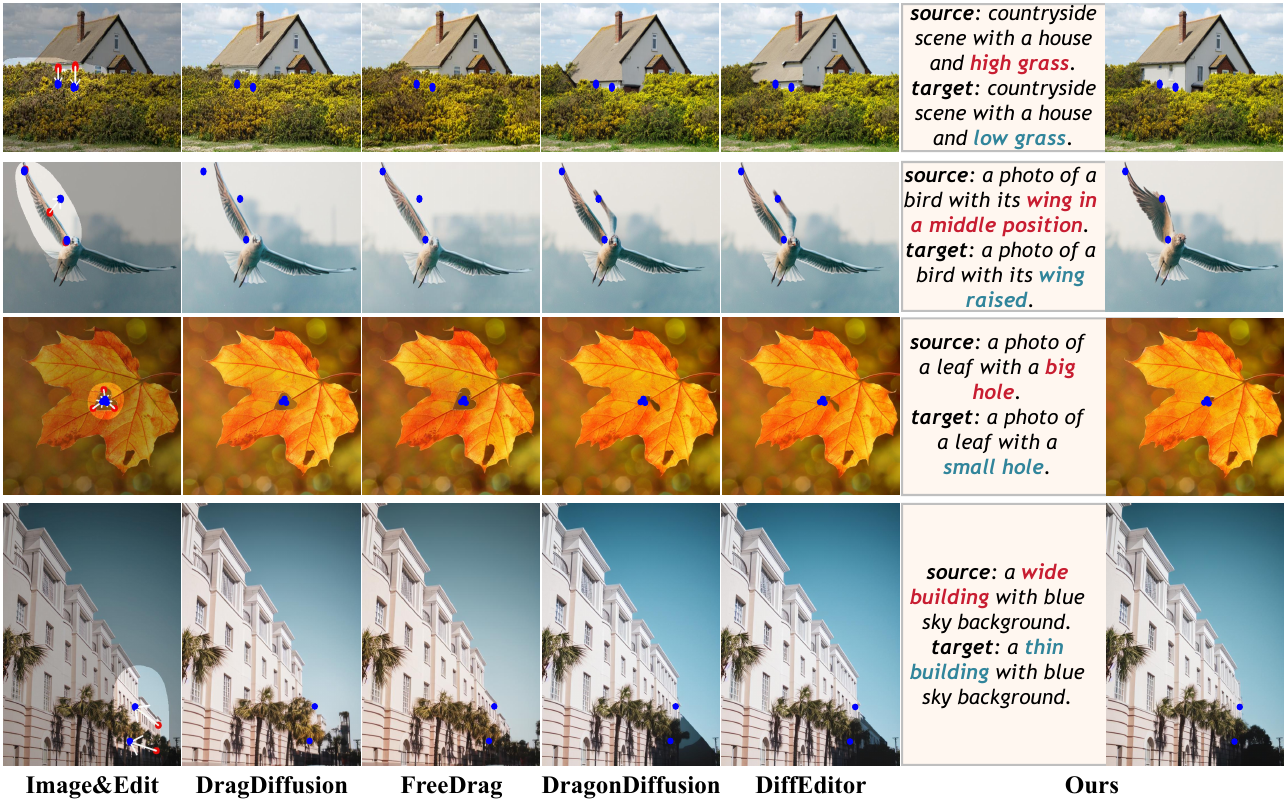}
    Qualitative comparison between our LucidDrag and other methods in drag-based editing.
    \label{fig:main-drag}
\end{figure*}

\renewcommand{\arraystretch}{1.1}{
\setlength{\tabcolsep}{1mm}{
\begin{table*}[t]
\centering
\caption{{Comparisons of content dragging on DragBench.}}
\vspace{1mm}
\resizebox{0.8\linewidth}{!}{%
\begin{tabular}{l|cccc|c}
\toprule
&DragDiffusion &FreeDrag  &DragonDiffusion &DiffEditor &Ours\\
\midrule
Mean Distance ($\downarrow$) &32.60 &30.40 &29.16 &$26.01\pm0.81$
&$\textbf{20.46}\pm\textbf{0.77}$\\
GScore ($\uparrow$) &6.94	&7.15	&6.26	&6.42	&\textbf{7.37}
 \\
\bottomrule
\end{tabular}}
\label{tab:detail_compare_md}
\end{table*}
}
}

\paragraph{Content dragging} We evaluate the proposed LucidDrag against existing drag editing models~\cite{ling2023freedrag, shi2023dragdiffusion, mou2023dragondiffusion,mou2024diffeditor}.
We first conduct quantitative comparisons.
Following DragDiffusion~\cite{shi2023dragdiffusion}, we utilize the DragBench benchmark which is designed for the image-dragging task. In DragBench, each image is accompanied by a set of dragging instructions, including several pairs of handle and target points and a mask indicating the editable region. For editing precision, we use the Mean Distance~\cite{pan2023drag} to evaluate the model's ability to move the contents to the target points. For image quality, existing Image Quality Assessment methods~\cite{ke2021musiq,zhang2018unreasonable} rely on handcrafted features or are trained on limited image samples, which do not always align well with human
perception~\cite{zhang2024gooddrag}. 
Thereby, we employ GScore~\cite{zhang2024gooddrag} to provide a human-aligned assessment of image quality via Large Multimodal Models.

The quantitative results are presented in Tab.~\ref{tab:detail_compare_md}. For dragging precision, our LucidDrag consistently outperforms other methods by a significant margin across all categories in Mean Distance, indicating higher accuracy in dragging handle contents to target positions.  In terms of image quality, our method achieves an average GScore of 7.37, surpassing DragDiffusion (6.94), FreeDrag (7.15), DragonDiffusion (6.26), and DiffEditor (6.42).

We also present qualitative results in Fig.~\ref{fig:main-drag}. DragDiffusion and FreeDrag has difficulty in accurately dragging corresponding contents to designated target locations. Although DragonDiffusion and DiffEditor can better recognize handle points and achieve more precise point movement, they tend to introduce artifacts, thereby reducing image quality. In contrast, LucidDrag demonstrates superior drag control by understanding potential intentions and providing semantic guidance. Furthermore, LucidDrag generates images with greater fidelity due to the explicit quality guidance it provides.

\paragraph{Object moving} We also conduct experiments on the object moving task, which can be seen as a special task of drag-style manipulation~\cite{mou2023dragondiffusion}. We compare our method with DragonDiffusion~\cite{mou2023dragondiffusion} and DiffEditor~\cite{mou2024diffeditor}. 
Following DragonDiffusion~\cite{mou2023dragondiffusion}, we select 20 editing samples as the test set. We calculate the CLIP distance between the edited results and the target description. Besides, inspired by~\cite{zhang2024gooddrag}, we utilize  Large Multimodal Models to evaluate the overall performance, denoted as the LMM score. The results are shown in Tab.~\ref{tab:comparison_moving}. Our approach achieves higher CLIP scores and LMM scores, demonstrating the promising performance of our method.

\begin{figure*}[hp]
  \centering
    \includegraphics[width=0.8\linewidth]{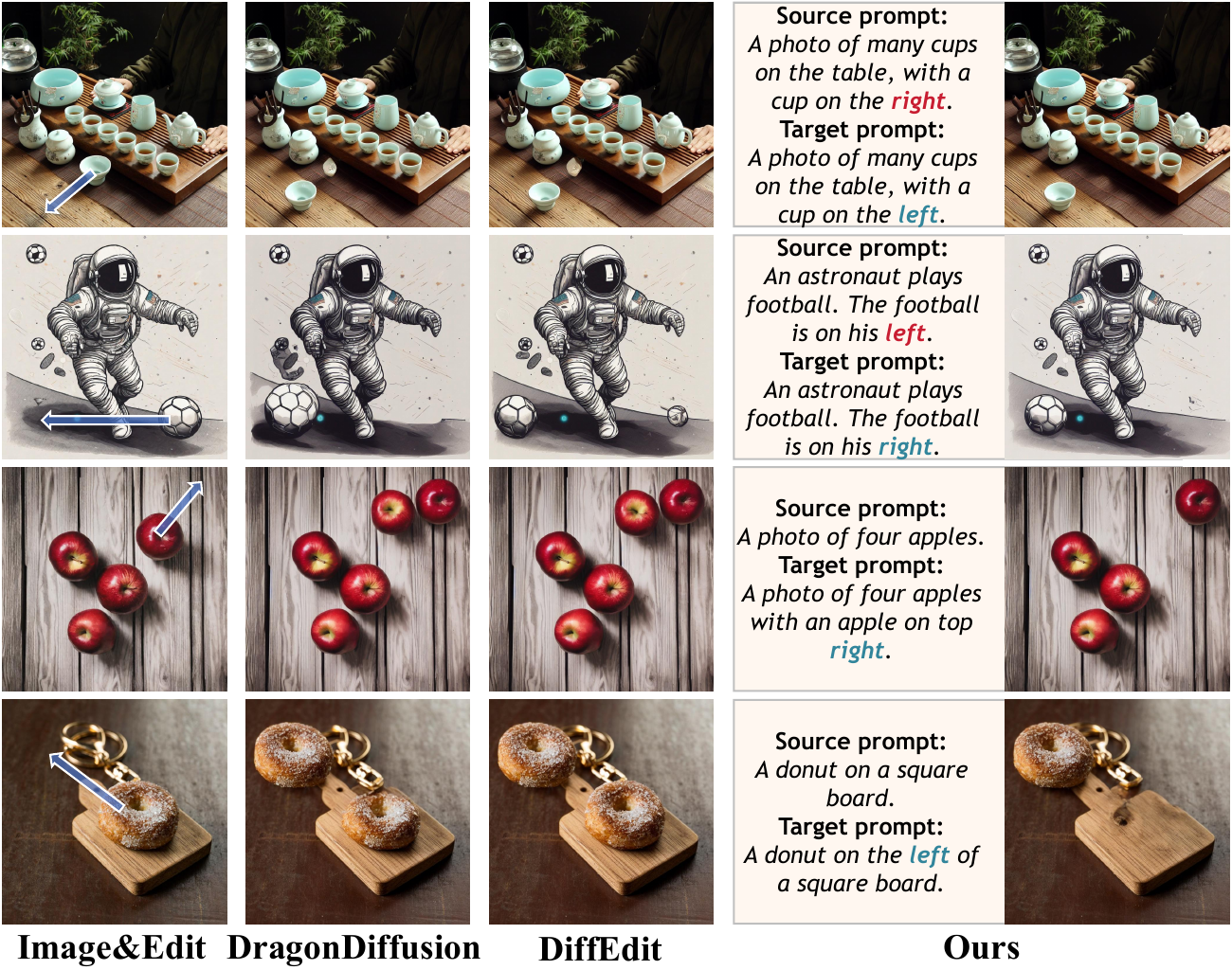}
    \caption{Qualitative comparison between our LucidDrag and other methods in object moving.}
    \label{fig:main-move}
\end{figure*}

Qualitative comparisons of the object moving task are shown in Fig.~\ref{fig:main-move}. Although the comparison methods can precisely recognize and generate objects in the target locations, our method shows better performance in image quality benefiting from our novel framework. On the one hand, the quality guidance constrains the distribution of the generated images, avoiding artifacts and unexpected changes during generation. For example, in the case of the water cup on the lower left, the human hands of comparison methods are deformed. Differently from prior work, our method achieves better image fidelity.
On the other hand, the intention predicted by our intention reasoner can provide semantic guidance, thus preventing the object from reappearing in the source location. For example, in the case of the doughnut on the lower right, the doughnut of comparison methods still appears in its original position.  Conversely, our method successfully moves the doughnut to the target position.

\renewcommand{\arraystretch}{1.1}{
\setlength{\tabcolsep}{2mm}{
\begin{table*}[ht]
\centering
\caption{{Compairisons of object moving.}}
\vspace{1mm}
\resizebox{0.5\linewidth}{!}{%
\begin{tabular}{l|cccc}
\toprule
&DragonDiffusion &DiffEdit &Ours\\
\midrule
CLIP-score ($\uparrow$)&0.255 & 0.257 &0.260\\
LMM-score ($\uparrow$)&6.25 &6.5 &7.25\\
\bottomrule
\end{tabular}}
\label{tab:comparison_moving}
\end{table*}
}
}

\subsection{Ablation Study}

\paragraph{Intention planner}
LucidDrag introduces the intention reasoner for intention understanding. In one respect, this module enhances semantic-aware capability during image editing. In another respect, it provides semantic guidance during generation, prompting accurate dragging of handle points to target positions. To substantiate our claims, we present an ablation study of the intention reasoner. 
We present qualitative experiments in Fig.~\ref{fig:main-ablation}.
For example, one possible semantic intention of the case in the first row of Fig.~\ref{fig:main-ablation} is to reduce the size of the wheel. When the intention reasoner is removed, \textit{i.e.}, \textit{w/o Intention}, the model has difficulty in understanding the semantic intention, limiting its dragging ability. Alternatively, our approach provides a strong semantic understanding of intentions, enabling precise dragging of objects to their target positions. Quantitative results are presented in Tab~\ref{tab:ablation}. Employing the Intention planner yields a 
$3.20\%$ performance improvement in Mean Distance and a $0.61\%$ performance improvement in GScore, further demonstrating the necessity of semantic understanding and the effectiveness of the intention planner.

\paragraph{Quality guidance}
One of the major differences between our LucidDrag and the previous ~\cite{mou2023dragondiffusion,pan2023drag,mou2024diffeditor,shi2023dragdiffusion} is that we explicitly introduce quality guidance to improve the image quality of the edited image. To verify the effectiveness of quality guidance, ablation studies are conducted both qualitatively and quantitatively. Specifically, we denote the ablation study of removing quality guidance as \textit{w/o Quality}. As depicted in Fig.~\ref{fig:main-ablation}, removing quality guidance leads to reduced image quality, evidenced by unexpected changes in the front wheel of the bicycle (the red iron bracket) and artifacts in the spinning top. 
Ablating reasoned intention reduces the model's ability to perceive the target shape. For example, after removing the intention reasoner, there are some distortions of the structure.
Additionally, Tab.~\ref{tab:ablation} demonstrates that ablating quality guidance not only degrades image quality but also impacts edit precision. This may because of the fact that the unexpected artifacts hinder the dragging performance. 

\makeatletter\def\@captype{figure}\makeatother
\begin{minipage}[m]{.45\linewidth}

  \centering
    \includegraphics[width=\linewidth]{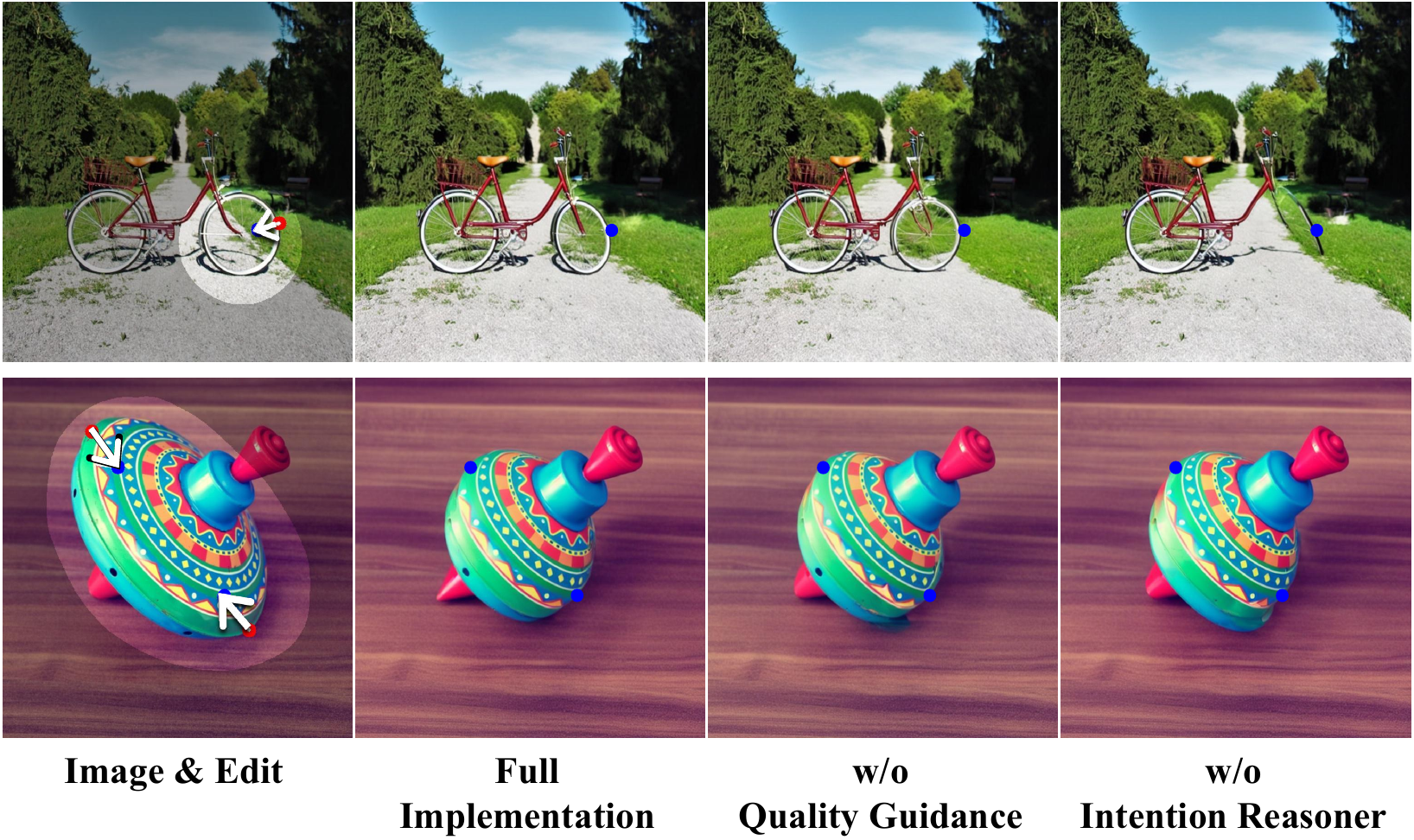}
    \vspace{-0.7em} 
    \caption{Visualization of ablation study.}
    \label{fig:main-ablation}
\end{minipage}
\hfill
\makeatletter\def\@captype{table}\makeatother
\begin{minipage}[b]{.55\linewidth}
\scriptsize
\renewcommand\arraystretch{1.1}
    \centering
    \caption{{Quantitative result of ablation study.}}
    \vspace{1mm}
    \resizebox{0.9\linewidth}{!}{%
    \begin{tabular}{l|cccc}
    \toprule
    &\makecell{Full Implementation\\(Ours)} &\makecell{w/o Intention \\ Reasoner} &\makecell{w/o Quality \\Guidance}\\
    \midrule
    Mean Distance ($\downarrow$) &$\textbf{20.46}\pm0.77$ &$23.66\pm0.73$ &$22.13\pm0.64$\\
    GScore ($\uparrow$) &\textbf{7.37} &6.76 &6.47\\
    \bottomrule
    \end{tabular}}
    \label{tab:ablation}
\end{minipage}

\section{Conclusion}
In this work, we identify the limitations of previous drag editing models in understanding semantic intentions and generating high-quality edited images. In response, we design a novel framework called LucidDrag, which involves an intention reasoner to clarify possible intentions and a collaborative guidance sampling mechanism that incorporates explicit semantic guidance and quality guidance.  
LucidDrag excels in: \textbf{i)} adequate understanding of semantic intention, improving semantic perception ability and diversity of the generated images; \textbf{ii)} enhanced dragging performance, including improved drag accuracy and image quality.  Extensive results show the efficiency of our approach and the potential for further advancements in semantic-aware drag-based editing.

\section{Limitations}
Although our method is capable of achieving semantic-aware drag-based editing without the need for training, there are some limitations. \textbf{i)} Complex objects are challenging to drag, and unexpected deformation can occur over long distances. This may be attributed to difficulties in comprehending the intricate nature of the object or inaccurate object tracking. In future work, we will investigate the potential for further improvements in performance by utilizing more powerful image generation models and incorporating comprehensive intention understanding. \textbf{ii)} Hyperparameters involved in the editing process will affect the editing results. In future work, we intend to utilize LLM as an agent to automatically determine model hyperparameters to enhance the editing performance. 

\noindent{\textbf{Acknowledgement}}
This research is sponsored by National Natural Science Foundation of China (Grant No. 62306041), Beijing Nova Program (Grant No. Z211100002121106, 20230484488), CCF-DiDi GAIA Collaborative Research Funds (Grant No. 202414), and Beijing Municipal Science \& Technology Commission (Z231100007423015).

\medskip

{\small
\bibliographystyle{ieee_fullname}
\bibliography{neurips_2024}
}
\newpage
\appendix

\section{Appendix} 
This appendix contains additional details for the NeurIPS 2024 submission, titled "Localize, Understand, Collaborate Guide: Semantic-aware Dragging via Intention Reasoner." The appendix is organized as follows:

\begin{itemize}
    \item \S\ref{Appendix:diffusion} Additional Preliminaries of Diffusion Model.
    \item \S\ref{Appendix:algorithm} Algorithm Pipeline of LucidDrag.
    \item \S\ref{Appendix:instruct} Instruct Prompts for LLM.
    \item \S\ref{Appendix:comparisions} More Comparsions.
    \item \S\ref{Appendix:analysis} More Analysis.
    \item \S\ref{Appendix:social} Social Impacts.
    
\end{itemize}

\subsection{Additional Preliminaries of Diffusion Model}\label{Appendix:diffusion}
\paragraph{DDIM}
In the inference time of a diffusion model, given an initial noise vector $z_T$, the noise is gradually removed by sequentially predicting the added noise for $T$ steps. DDIM ~\cite{song2020denoising} is one of the efficient denoising approaches that follow a deterministic process, in contrast to the original stochastic diffusion process:
\begin{equation}
\label{eq:ddim2}
    z_{t-1}=\sqrt{\frac{\alpha_{t-1}}{\alpha_{t}}}z_{t}+\left(\sqrt{\frac{1}{\alpha_{t-1}}-1}-\sqrt{\frac{1}{\alpha_{t}}-1}\right)\cdot\tilde{\varepsilon}_\theta,
\end{equation}
where $\tilde{\varepsilon}_\theta$ is the estimated noise.

\textbf{DDIM inversion ~\cite{song2020denoising,mokady2023null} refers to a technique that aims to transform an input image into a noise vector $z_T$ conditioned on a given prompt or target representation. This process is accomplished by reversing the diffusion process, whereby the final noise sample $z_T$ is returned to the initial $z_0$.}
\begin{equation}
\label{eq:ddim_inverion}
    z_{t+1}=\sqrt{\frac{\alpha_{t+1}}{\alpha_{t}}}z_{t}+\left(\sqrt{\frac{1}{\alpha_{t+1}}-1}-\sqrt{\frac{1}{\alpha_{t}}-1}\right)\cdot\tilde{\varepsilon}_\theta.
\end{equation}

\paragraph{Classifier-free guidance}
For text-based image generation using diffusion models, the classifier-free guidance technique is often employed to address the challenge of amplifying the effect of the text condition. To this end, Ho et al.~\cite{ho2021classifier} have presented the classifier-free guidance technique, whereby the prediction is also performed unconditionally and then extrapolated with the conditioned prediction. Specifically, the estimated noise in Eq.~\ref{eq:pred_noise} is adjusted as follows:
\begin{equation}
\tilde{\varepsilon}_\theta(z_t,t,y,\varnothing)=\varepsilon_\theta(z_t,t,\varnothing) \\
 +w\cdot(\varepsilon_\theta(z_t,t,y)-\varepsilon_\theta(z_t,t,\varnothing)),\\
\end{equation}
where $\varnothing=\psi($""$)$ is the embedding of a null text.
$\varepsilon_\theta(z_t,t,y)$ represents the conditional predictions.
$w$ is the guidance scale parameter.

\paragraph{Classifier guidance}

The diffusion sampling process can be guided by a variety of energy functions, $g(z_t; t, y)$, which are not limited to probabilities derived from a classifier. Such energy functions may comprise, for instance, the approximate energy derived from another model \cite{liu2022compositional}, a similarity score derived from a CLIP model \cite{nichol2022glide}, time-independent energy in universal guidance \cite{bansal2023universal}, bounding box penalties on attention \cite{chen2024training}, or any attributes of the noisy images. 

The combination of this additional guidance with  \textit{"classifier-free guidance"}\cite{ho2021classifier} enables the generation of high-quality text-to-image samples that also possess low energy according to the energy function $g$:
\begin{equation}
\hat\eps_t = (1 + s) \eps_\theta(z_t; t, y) - s \eps_\theta(z_t; t, \emptyset)  + \eta \nabla_{z_t} g(z_t; t, y),
\label{eq:propmatch}
\end{equation}
where $s$ represents the strength of the classifier-free guidance; $v$ is an additional weight for the guidance provided by $g$. As with classifier guidance, we scale by $\sigma_t$ to convert the score function to a prediction of $\eps_t$. Our work contributes by identifying energy functions $g$ that can be used to control the properties of objects and interactions between them.

\subsection{Algorithm Pipeline of LucidDrag} \label{Appendix:algorithm}
To facilitate the understanding of our LucidDrag, we present the entire algorithm pipeline in Algorithm~\ref{aug}.  The intention reasoner employs a locator and a reasoner to infer the underlying intentions. Subsequently, collaborative guidance is leveraged to generate images through the integration of semantic guidance, quality guidance, and editing guidance.

\begin{algorithm}[ht]
\caption{Proposed LucidDrag}
\label{aug}
\textbf{Require}:

\hspace{0.01cm}pre-trained SD $\epsilon_{\boldsymbol{\theta}}$; image to be edited ${x}_{0}$; editing guidance steps $n_1$. quality guidance steps $n_2$ ($n_2<n_1$).

\textbf{Intention Reasoner}:

    \hspace{0.01cm} (1) Locatize the interested region, $O=LVLM(x, P)$.
    
    \hspace{0.01cm} (2) Understand the possible intentions, $D=LLM(O,C,P)$.

    \hspace{0.01cm} (3) Sample source prompt and target prompt: $p_{j}^{s}, p_{j}^{t}$.
    
\textbf{Collaborative Guidance}:

    \hspace{0.01cm} (1) ${z}_0=Encoder({x}_0)$
    \hspace{0.01cm} (2) Invert ${z}_0$ to ${z}_{T}^{gud}$, ${z}_{t+1}^{gud}=\mathit{DDIM\_inversion}({z}_{t}^{gud},p_{*}^{s})$. Then, build the memory bank.
    
     \hspace{0.01cm} (3) Initialize ${z}_{T}^{gen}$ with ${z}_{T}^{gud}$.
     
 \For{$t=T,\ \ldots,\ 1$}{
noise prediction: $\hat{\boldsymbol{\epsilon}}_t=\epsilon_{\boldsymbol{\theta}}({z}_{t}, t,{c},{c}_{im})$;

\If{$T-t<n_1$ }{compute $g_{edit}$ by Eq.~\ref{eq_edit};

\eIf{$T-t<n_2$ }{compute $g_{guality}$ by Eq.~\ref{eq_gquality};

compute energy function by Eq.~\ref{eq_collabguid};
}{
set energy function as $g_{edit}$;
}

inject gradient guidance by Eq.~\ref{eq_classifierguid};

}

compute ${z}_{t-1}$ by Eq.~\ref{eq:ddim};
 }

\hspace{0.01cm} ${x}_0 = Decoder({z}_0)$;

\textbf{Output:} ${x}_0$
\end{algorithm}

\subsection{Instruct Prompts for LLM}\label{Appendix:instruct}
Our approach employs the advanced spatial and reasoning capabilities of LLMs to infer potential intentions. The specific instructions provided in this work are presented below.

\rule{\linewidth}{0.3mm}
\noindent \textit{\textbf{Instruct prompt for reasoning potential intentions. }}\\
\texttt{
You are a helpful assistant. Given the original description and human-select drag information (including the description of the start point and the drag direction),
reason the intention of the human based on the drag information, then generate a refined source prompt and target prompt. The intention of the human is dragging some part of the image to make some deformation (bigger, shorter, longer, open, closed, etc.) or change its posture (looking left, looking right, looking sideways, looking closer, etc.).
The source prompt and target prompt should be similar and reflect the difference before and after dragging. Notably, if the difference between the source and target prompt is hard to describe, you can directly set the source prompt and target prompt as the original description. For example:\\
example 1: \\
INPUT: \\
Original description: a red motorcycle. 
Drag information 1: 
The start point is the behind of the motorcycle. 
Direction is [[292, 276], [345, 275]]. \\
OUTPUT: \\
Intention: making the motorcycle longer.\\
Source prompt: a short red motorcycle.\\
Target prompt: a long red motorcycle.\\
example 2: \\
INPUT: \\
Original description: a photo of a raccoon. drag information 1: The start point is the nose of a squirrel. Direction is [[297, 270], [347, 248]].\\
OUTPUT:\\ 
Intention: make the raccoon look sideways.\\
Source prompt: a photo of a raccoon looking forward.\\
Target prompt: a photo of a raccoon looking sideways.\\
...\\
}
\rule{\linewidth}{0.3mm}

\subsection{More Comparisons}\label{Appendix:comparisions}
\paragraph{Quantitative comparison}
In the primary paper, Tab.~\ref{tab:detail_compare_md} presents the Mean Distance and Gscore, averaged over all samples in the DragBench dataset. This supplementary section presents a detailed comparison of the results obtained for each individual category within the DragBench dataset.

The comparison in terms of Mean Distance is presented in Tab.~\ref{tab_sup:detail_compare_md}, while the comparison in terms of GScore is provided in Tab.~\ref{tab_sup:detail_compare_if}. 
As demonstrated in Tab.~\ref{tab_sup:detail_compare_md}, the proposed LucidDrag method achieves the lowest mean distance, thereby confirming the superiority of our technique in accurately dragging the content to the target position.

About the image fidelity metric as reported in Tab.~\ref{tab_sup:detail_compare_if}, our LucidDrag approach also demonstrates superior overall average performance. While methods such as DragDiffusion and FreeDrag achieve higher G-scores on some individual classes, this is likely due to their use of additional model fine-tuning to better fit the input image, which can be time-consuming.

Despite comparable image fidelity, the competing methods are deficient in terms of dragging accuracy, as evidenced by the elevated Mean Distance values in Tab.~\ref{tab_sup:detail_compare_md}. This demonstrates the superiority of the LucidDrag method in maintaining both high-quality image generation and precise content manipulation capabilities.

\renewcommand{\arraystretch}{1.2}{
\setlength{\tabcolsep}{0.6mm}{
\begin{table*}[ht]
\centering
\caption{\textbf{Comparisons of Dragging Accuracy (Mean Distance) on {DragBench} ($\downarrow$).}}
\vspace{2mm}
\resizebox{\linewidth}{!}{%
\begin{tabular}{l|cccccccccc|c}
\toprule
&Artworks &Landscape &City & Countryside &Animals & Head &Upper body &Full body &Interior &Other &Average \\
\midrule
DragDiffusion & 30.74 & 36.55 & 27.28 & 43.21 & 39.22 & 36.43 & 39.75 & 20.56 & 24.83 & 39.52 &32.60\\
FreeDrag &31.16	&31.92	&28.87	&33.57	&32.12	&39.02	&34.46	&19.75	&23.27	&36.11	&30.40\\
DragonDiffusion &26.23	&25.42	&29.23	&34.17	&36.13	&28.86	&48.61	&6.97	&18.63	&37.36	&29.16\\
DiffEditor &$24.41$	&$27.52$	&$34.94$	&$38.42$	&$25.62$	&$24.60$	&$24.77$	&$6.81$	&$17.71$	&$35.32$	&$26.01$
\\
\midrule
Ours &$\textbf{19.71}$	 &$\textbf{17.76}$	&$\textbf{26.74}$	&$\textbf{26.67}$	&$\textbf{22.74}$	&$\textbf{23.87}$	&$\textbf{20.70}$	&$\textbf{5.56}$	&$\textbf{14.34}$	&$\textbf{26.54}$	&$\textbf{20.46}$

\\
\bottomrule
\end{tabular}}

\label{tab_sup:detail_compare_md}
\end{table*}
}
}

\renewcommand{\arraystretch}{1.2}{
\setlength{\tabcolsep}{0.6mm}{
\begin{table*}[ht]
\centering
\caption{\textbf{Comparisons of Image Fidelity (GScore) on { DragBench} ($\uparrow$).}}
\vspace{2mm}
\resizebox{\linewidth}{!}{%
\begin{tabular}{l|cccccccccc|c}
\toprule
&Artworks &Landscape &City & Countryside &Animals & Head &Upper body &Full body &Interior &Other &Average \\
\midrule
DragDiffusion &6.71	&6.19	&\underline{7.19}	&\textbf{6.91}	&\underline{7.10}		&7.70 &\underline{7.70}	&7.30	&7.15	&6.74	&6.94\\

FreeDrag &\underline{7.04}	&\textbf{7.24}	&6.88	&6.37	&7.06	&\textbf{8.28} &\textbf{8.18}	&\underline{7.33}	&\underline{7.49}	&\underline{7.07}	&\underline{7.15}\\

DragonDiffusion &6.09	&6.07	&6.12	&5.47	&6.58	&6.93	&6.42 &6.93	&7.03	&6.09	&6.26\\

DiffEditor &6.90	&6.58	&6.656	&6.58	&6.47	&6.22	&6.32	&6.52	&6.78	&6.58	&6.62\\
\midrule
Ours &\textbf{7.28}	&\underline{6.95}	&\textbf{7.28}	&\underline{6.81}	&\textbf{7.14}	&\underline{7.75}	&7.54	&\textbf{8.05}	&\textbf{8.16}	&\textbf{7.43}	&\textbf{7.37}\\
\bottomrule
\end{tabular}}
\label{tab_sup:detail_compare_if}
\end{table*}
}
}

\begin{figure*}
  \centering
    \includegraphics[width=\linewidth]{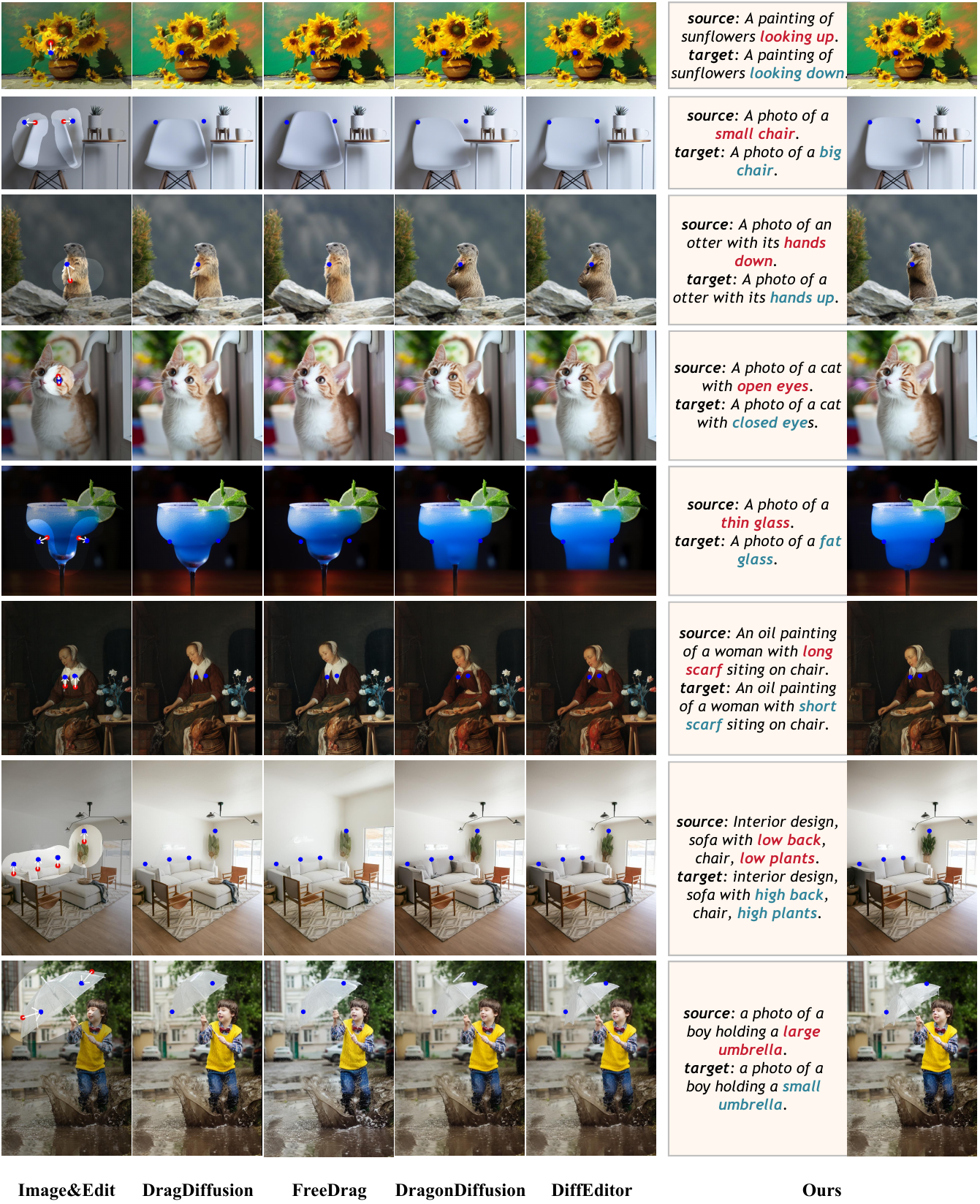}
    \vspace{-0.7em} 
    \caption{More qualitative comparison on content dragging.}
    \label{fig:drag_appendix}
\end{figure*}

\paragraph{More Qualitative Results}
More comparison results for the content dragging task are presented in Fig.~\ref{fig:drag_appendix}. These results, in conjunction with those presented in Fig.~\ref{fig:main-drag} of the main paper, substantiate the superiority of our method in terms of both drag-based editing performance and image quality maintenance, and also demonstrate the generality of our approach.

\begin{figure*}
\centering
\includegraphics[width=\linewidth]{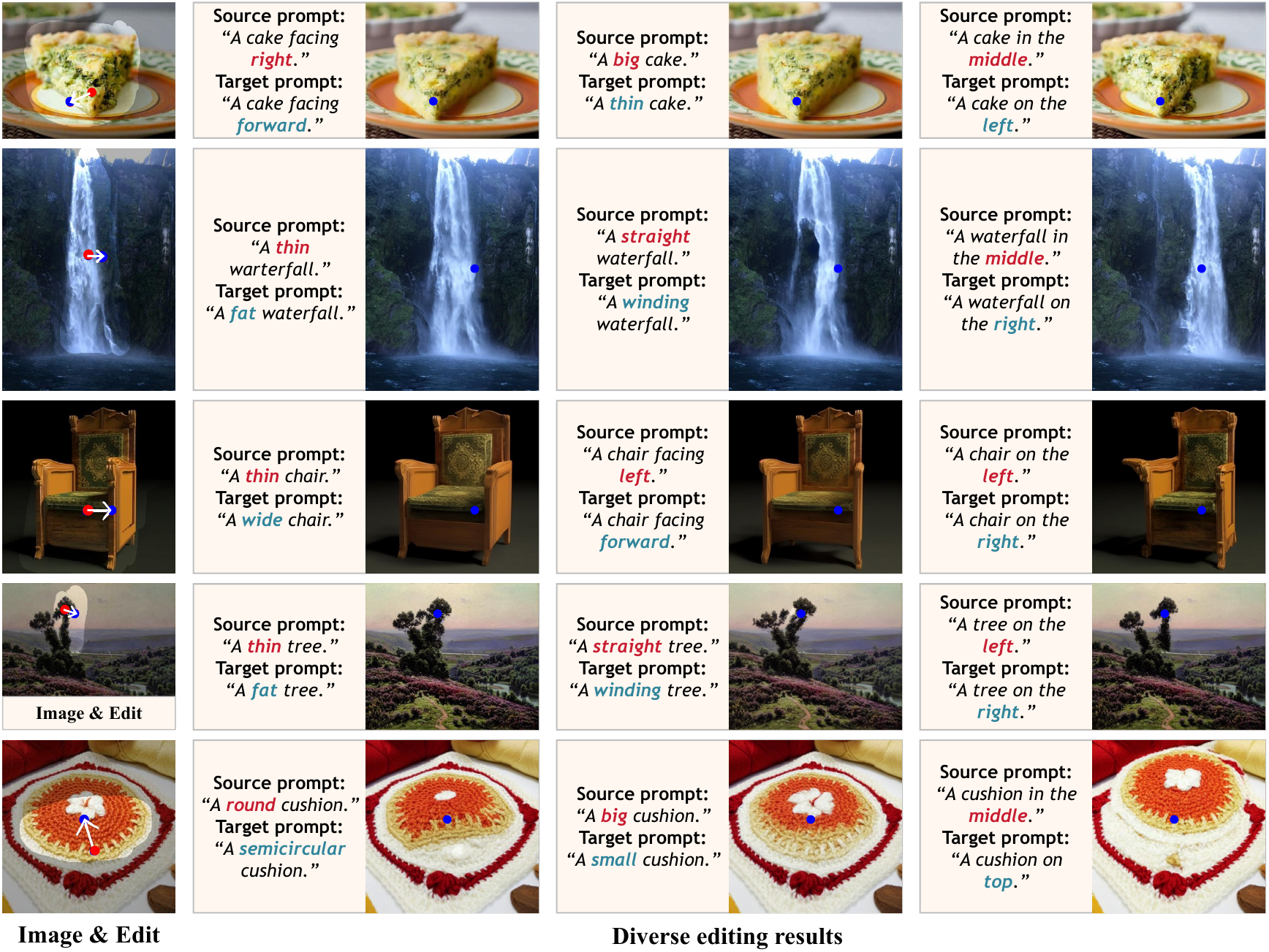}
\vspace{-0.7em}
\caption{LucidDrag allows generating diverse results.}
\label{fig:append-diversity}
\end{figure*}

Subsequently, we present additional qualitative results that demonstrate the semantic awareness capability of our proposed method. As illustrated in Fig.~\ref{fig:append-diversity}, our approach is capable of generating a diverse range of output images that are highly faithful to the intentions that have been deduced.

\subsection{More Analysis}\label{Appendix:analysis}
\paragraph{Analysis of different LVLMs and LLMs in Intention Reasoner}
We conduct experiments to examine the performance of different LVLMs and LLMs in the Intention Reasoner module. Specifically, we utilize Osprey~\cite{yuan2023osprey} and Ferret~\cite{you2023ferret} for LVLM and Vicuna~\cite{chiang2023vicuna}, LLama3~\cite{touvron2023llama}, and GPT 3.5~\cite{chatgpt2022} for LLM. We test various combinations, with Osprey+GPT3.5 being the default setting in our paper. As shown in Table~\ref{tab:r1}, all combinations outperform the experiment without the Intention Reasoner, confirming its reliability. This reliability stems two-fold: the LVLMs are trained with large-scale point-level labeled data and can easily achieve point-level understanding~\cite{yuan2023osprey}. Therefore, they can understand the user-given points without further fine-tuning. For the LLMs, state-of-the-art LLMs have been proven to possess strong spatial reasoning abilities~\cite{gurneelanguage}, enabling them to deduce reasonable intentions.

\renewcommand{\arraystretch}{1.1}{
\setlength{\tabcolsep}{2mm}{
\begin{table*}[ht]
\centering
    \caption{Results with different LVLMs and LLMs} 
    \vspace{1mm}
    \resizebox{\linewidth}{!}{%
    \label{tab:r1} 
    \begin{tabular}{lccccccc}
        \toprule
        & \textbf{\makecell{w/o Intention\\ Reasoner}} & \textbf{\makecell{Ferret+\\Vicuna}} & \textbf{\makecell{Ferret+\\LLama3}} & \textbf{\makecell{Ferret+\\GPT3.5}} & \textbf{\makecell{Osprey+\\Vicuna}} & \textbf{\makecell{Osprey+\\LLama3}} & \textbf{\makecell{Osprey+\\GPT3.5 (Ours)}} \\
        \midrule
        Mean Distance (\(\downarrow\)) & 23.66 & 22.49 & 21.96 & 20.65 & 20.84 & 20.48 & \textbf{20.46} \\
        GScore (\(\uparrow\))          & 6.76  & 7.12  & 7.11  & 7.35  & 7.27  & 7.13  & \textbf{7.37} \\
        \bottomrule
    \end{tabular}}
\end{table*}
}}

\begin{figure*}[h]
\centering
\includegraphics[width=\linewidth]{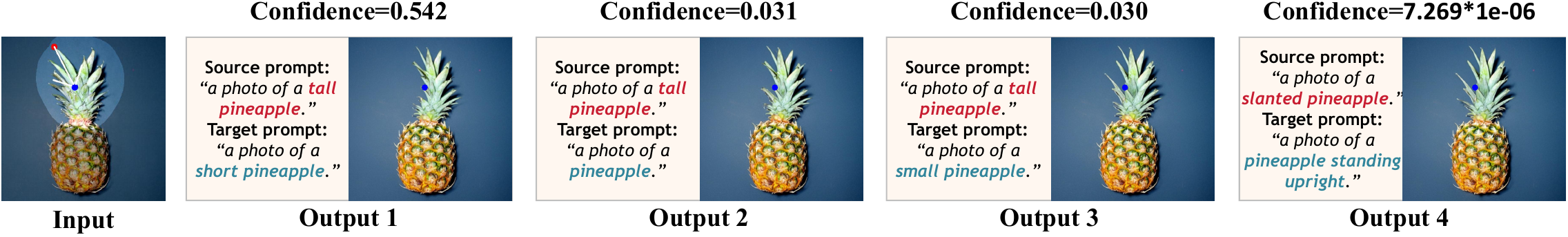}
\vspace{-0.7em}
\caption{Analysis of confidence probabilities. A higher confidence probability indicates that the intention of the output is more reasonable, leading to better editing results.}
\label{fig:confidence}
\end{figure*}

\paragraph{Analysis of the confidence probabilities}
We analyze the confidence probabilities of the intention reasoner. As shown in Fig.~\ref{fig:confidence}, the confidence probability reflects the quality of the output text in LLM. A higher confidence probability indicates that the intention of the output is more reasonable, leading to better editing results.

\begin{figure*}[h]
\centering
\includegraphics[width=\linewidth]{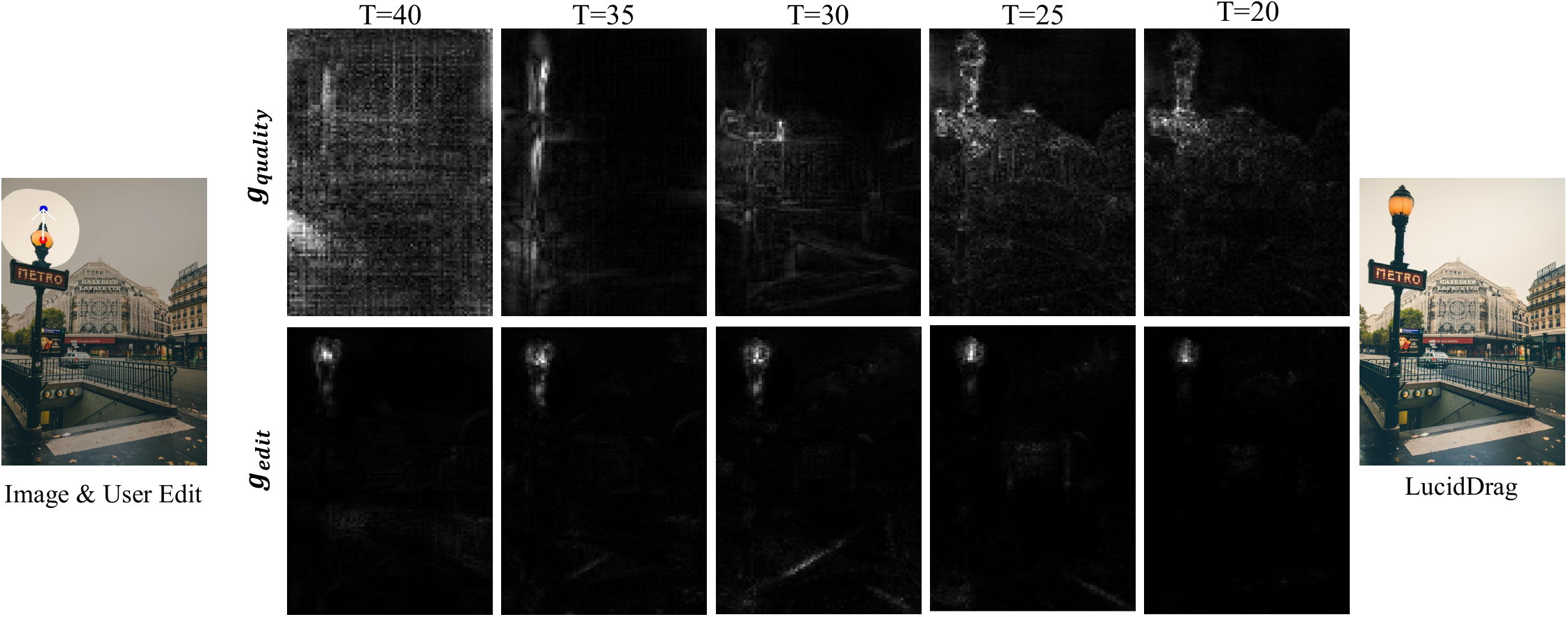}
\vspace{-0.7em}
\caption{Visualization of the quality guidance and editing guidance.}
\label{fig:quality-guidance}
\end{figure*}

\paragraph{Analysis of the guidance}
Fig.~\ref{fig:quality-guidance} illustrates the evolution of gradient maps during the drag-based editing task at different time steps. The first row of the figure depicts the gradient maps produced by the editing guidance, $g_{edit}$, while the second row depicts the gradient maps generated by the quality guidance, $g_{quality}$. The visualizations presented demonstrate a process of gradual convergence. In particular, as the sampling progresses, the activation range of the gradient maps narrows progressively, gradually converging toward the respective editing areas.

\begin{figure*}[h]
\centering
\includegraphics[width=0.9\linewidth]{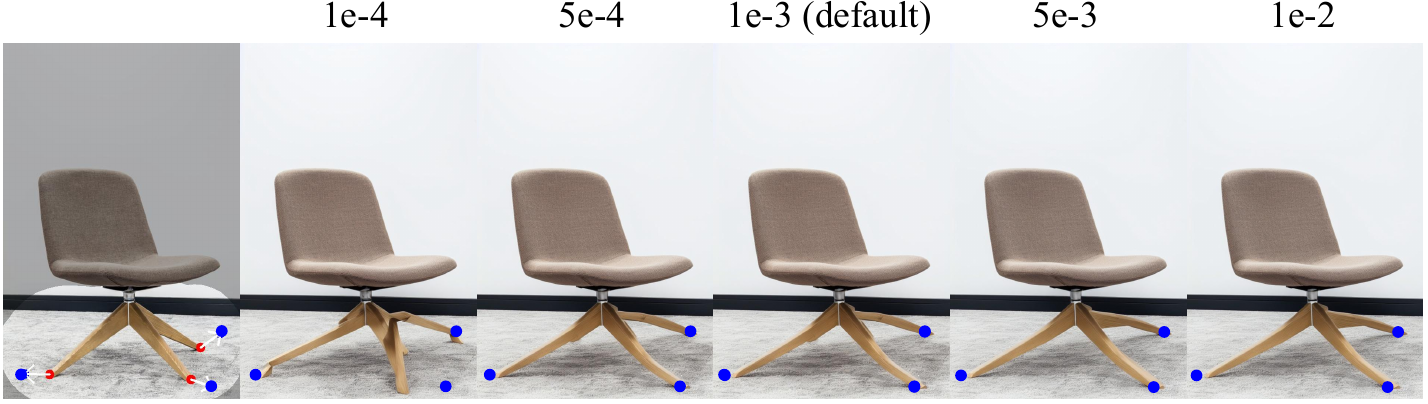}
\vspace{-0.7em}
\caption{Analysis of the quality guidance weight.}
\label{fig:Hyperparameters}
\end{figure*}

Fig.~\ref{fig:Hyperparameters} presents a visual analysis of the influence of the weight of quality guidance $w_{quality}$ on the image editing results. 
The figure illustrates that increasing the weights can strengthen the importance of their respective energy functions. However, as a greater degree of editing is more likely to result in distortion or artifacts in the image, there is a trade-off between the editing effects and the image quality. 
Consequently, in our design, we have set $w_{quality}$ as 1e-3 to avoid excessive constraints from the quality guidance, which could otherwise limit the effectiveness of the image editing process.

\paragraph{The efficiency of different methods}
We present the efficiency of different methods in Table~\ref{tab:r2}. Our method has a relatively small inference time and comparable memory requirements.
\begin{table}[h]
    \centering
    \caption{Efficiency of different methods.} 
    \label{tab:r2} 
    \begin{tabular}{lccccc}
        \toprule
        & \textbf{DragDiffusion} & \textbf{FreeDrag} & \textbf{DragonDiffusion} & \textbf{DiffEditor} & \textbf{Ours} \\
        \midrule
        Time (s) \(\downarrow\) & 80 & 92 & 30 & 35 & 48 \\
        Memory (GB) \(\downarrow\) & 12.8 & 13.1 & 15.7 & 15.7 & 15.8 \\
        \bottomrule
    \end{tabular}
\end{table}

\begin{figure*}[h]
  \centering
    \includegraphics[width=0.8\linewidth]{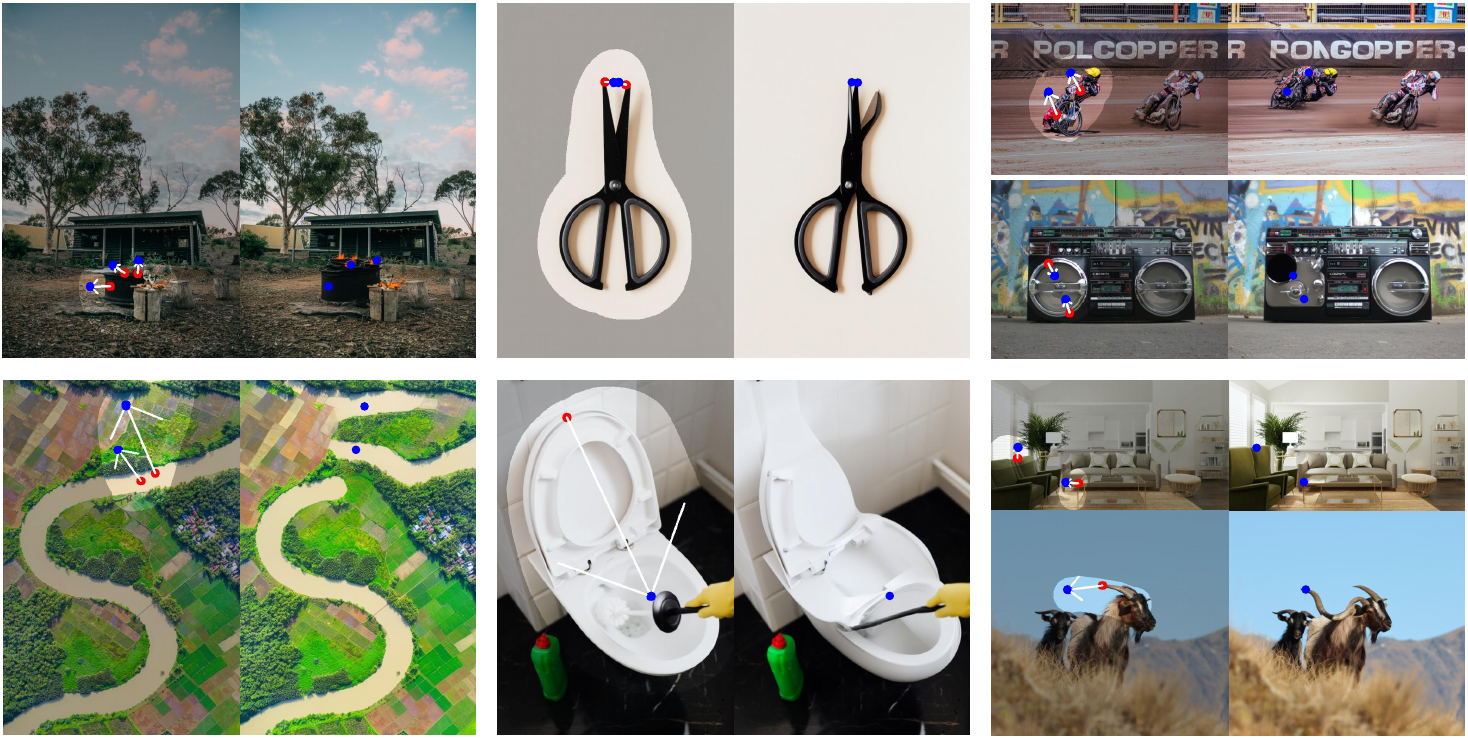}
    \vspace{-0.7em} 
    \caption{Bad case of our LucidDrag.}
    \label{fig:bad_case}
\end{figure*}

\paragraph{Bad case}
As illustrated in Fig.~\ref{fig:bad_case}, some complex objects are challenging to drag and some unexpected deformation occurs when dragging over long distances. This may be attributed to difficulties in comprehending the intricate nature of the object or inaccurate object tracking. 

\subsection{Social Impacts}\label{Appendix:social}
The emerging LucidDrag technology has promising applications in image editing, content creation, and visual design, inspiring people to create art. However, the use of this technology requires a critical evaluation of potential negative consequences, such as the generation of false or misleading content and potential privacy violations. To address these concerns, the integration of robust illegal content identification models is a viable approach to mitigate these risks and ensure the responsible and ethical use of LucidDrag within the creative landscape.

\end{document}